\documentclass[10pt,twocolumn,a4paper]{esaAI}

\usepackage{glossaries}
\usepackage{graphicx}
\usepackage{subcaption}
\usepackage{xcolor}

\newacronym{lsatc}{LSatC}{Low Earth Orbit Satellite Constellation}
\newacronym{ma-drl}{MA-DRL}{Multi-Agent Deep Reinforcement Learning}
\newacronym{fl}{FL}{Federated Learning}
\newacronym{e2e}{E2E}{end-to-end}
\newacronym{isl}{ISL}{inter-satellite link}
\newacronym{gsl}{GSL}{ground-to-satellite link}
\newacronym{snr}{SNR}{signal-to-noise ratio}
\newacronym{fifo}{FIFO}{first-in first-out}
\newacronym{dnn}{DNN}{Deep Neural Network}
\newacronym{ddqn}{DDQN}{Double Deep Q-Learning}
\newacronym{rl}{RL}{Reinforcement Learning}
\newacronym{cka}{CKA}{Centered Kernel Alignment}
\newacronym{cdf}{CDF}{Cumulative Distribution Function}
\newacronym{sfl}{SFL}{Satellite Federated Learning}

\newcommand{\flc}[1]{#1}

\title{An open source Multi-Agent Deep Reinforcement Learning Routing Simulator for satellite networks}

\author{Federico Lozano-Cuadra, Mathias D. Thorsager, Israel Leyva-Mayorga, and Beatriz Soret
\thanks{\makebox[0pt][l]{\textsuperscript{}}F. Lozano-Cuadra (flozano@ic.uma.es) and B. Soret are with the Telecommunications Research Institute, University of Malaga, 29071, Malaga, Spain. M. D. Thorsager and I. Leyva-Mayorga are with the Connectivity Section, Aalborg University, 9220, Aalborg, Denmark. The work of F. Lozano-Cuadra and B. Soret is partially funded by the Spanish Ministerio de Ciencia, Innovación y Universidades (“TATOOINE”, PID2022-136269OB-I00) and by ESA SatNEx V (prime contract no. 4000130962/20/NL/NL/FE). The view expressed herein can in no way be taken to reflect the official opinion of the European Space Agency (ESA). 

This paper was accepted ans presented at the SPAICE conference, organized by ESA. For more information, visit https://spaice.esa.int/.
}
}

\begin{document}

\makeCustomtitle
\makeatletter
\let\@fnsymbol\@arabic
\makeatother


\begin{abstract}
This paper introduces an open source simulator for packet routing in \glspl{lsatc}. The simulator, implemented in \emph{Python}, supports traditional Dijkstra's based routing as well as more advanced learning solutions based on Q-Routing and \gls{ma-drl} \flc{from our previous work}. It uses an event-based approach with the \emph{SimPy} module to accurately simulate packet creation, routing and queuing, providing real-time tracking of queues and latency. The simulator is highly configurable, allowing adjustments in routing policies, traffic, ground and space segment topologies, communication parameters, and learning hyperparameters. 
Key features include the ability to visualize system motion and track packet paths while  considering the inherent uncertainties of such a dynamic system. Results highlight significant improvements in \gls{e2e} latency using \gls{rl}-based routing policies compared to traditional methods. 
The source code, the documentation and a \emph{Jupyter notebook} with post-processing results and analysis are available on \emph{GitHub}.
\end{abstract}



\section{Introduction}

Efficient routing in \glspl{lsatc} is critical for global connectivity in 6G networks. This requires addressing multiple challenges, including the partial knowledge of the network at the satellites and their continuous movement, and the time-varying sources of uncertainty in the system, such as traffic, communication links, or communication buffers~\cite{lozanocuadra2024continual}. Traditional routing algorithms are inadequate to address these problems: They either lack adaptability to network changes or congest the network with feedback messages. 
To overcome these challenges, new algorithms must be developed, some of them \gls{rl}-based, which need to be accompanied by a robust framework. \emph{Python} is the best environment for developing \gls{rl}-based algorithms due to its extensive libraries for machine learning, such as \emph{Keras-TensorFlow}, \emph{PyTorch}, \emph{NumPy}, and \emph{Pandas}. This paper introduces an open source \gls{ma-drl} Routing Simulator for satellite networks  built in \emph{Python}, where these designed algorithms can be implemented and tested. The simulator supports various routing algorithms, including some Dijkstra's~\cite{dijkstra1959} shortest path-based and those from our recent works: (1) The Q-Routing with Q-tables for distributed routing decisions~\cite{soret2023qlearning}, and (2) the \gls{ma-drl} first proposed in~\cite{lozanocuadra2024multiagent}, which was then further tested and extended to continual learning with \gls{sfl} in~\cite{lozanocuadra2024continual}. The source code, a \emph{Jupyter notebook} with some post-processing results and analysis, and the documentation for the \gls{ma-drl} Routing Simulator are available on \emph{GitHub}~\cite{lozano2024madrl}.


\begin{figure}[t]
   \centering
   \includegraphics[width=.95\columnwidth]{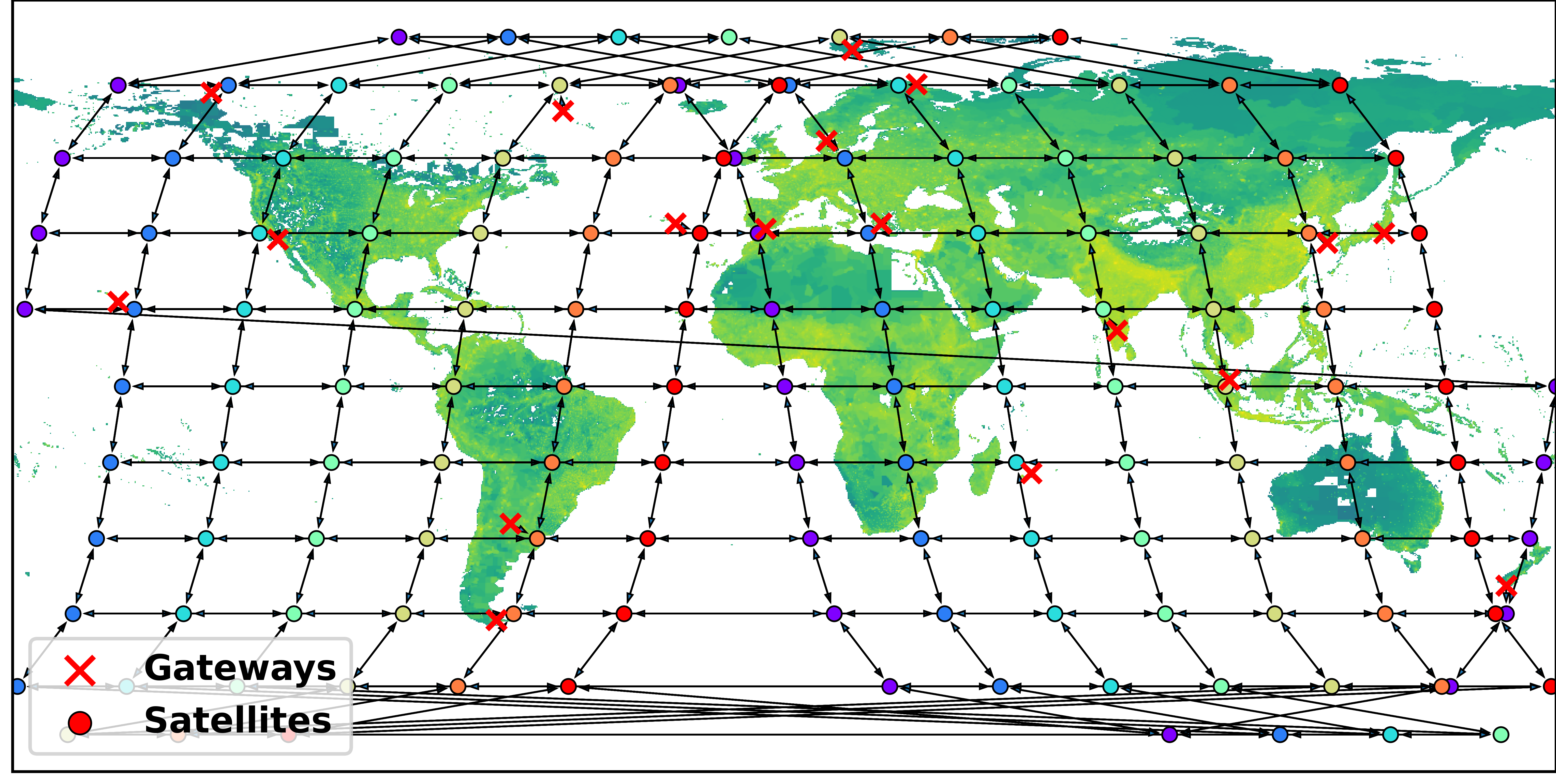}
   \caption{Kepler constellation deployed and their corresponding \glspl{isl} established following the \emph{Greedy matching} with 18 active gateways over the population maps~\cite{CIESIN}, \flc{where the green tone depends on the population density}. Each \flc{satellite's} colour is a different orbital plane.}
   \label{fig:ISLs_Kepler}\vspace{-0.5cm}
\end{figure}


\begin{figure*}[t]
    \centering
    \includegraphics[width=.95\textwidth]{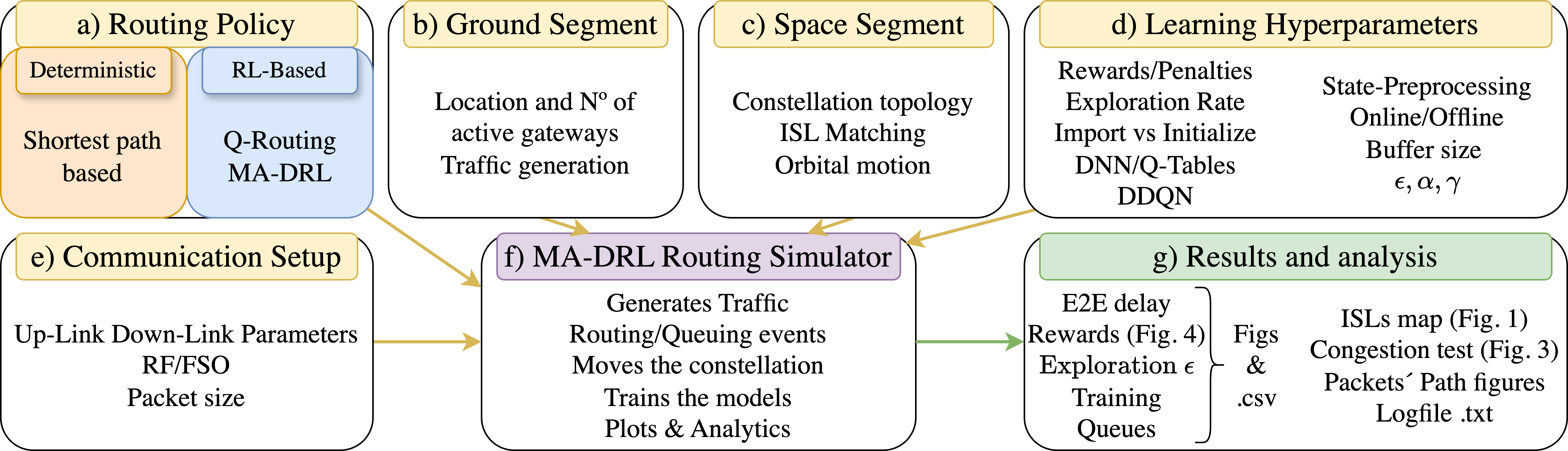}
	\caption{Input-Output \gls{ma-drl} Routing Simulator workflow.}
	\label{fig:system}
\end{figure*}


\section{Simulator architecture}

The event-based simulation environment was developed in \emph{Python} using the \emph{SimPy} module, chosen for its effectiveness in discrete event modeling~\cite{zinoviev2024discrete}. Time in the simulator progresses by jumping from one scheduled event to the next, rather than continuously. Each action, including creating, routing, and queueing of individual data packets, is explicitly simulated as a \emph{SimPy} event, providing accurate real-time tracking of queues and latency. Packets are unique entities (\flc{objects}) existing from creation at a generating gateway until they arrive at the destination gateway. The transmission time is calculated based on the packet size and current link rates, propagation time based on the exact distance between transmitter and receiver at transmission time, and queue time based on the time a packet spends in the queue. This detailed level of simulation is essential for representing the states and computing the rewards of the environment in our \gls{rl}-based routing algorithms.

The simulator emulates a realistic scenario where ground gateways gather the nearby terrestrial traffic that is assumed to be generated by mobile users and distribute that to each other gateway equally through a \gls{lsatc}, integrating space and ground segments into the communication network, as shown in Fig.~\ref{fig:ISLs_Kepler}. The environment is built as a \flc{time-variant} dynamic graph $\mathcal{G}\flc{_t}(\mathcal{N}, \mathcal{E})$ with nodes $\mathcal{N}$, representing satellites and gateways, and edges $\mathcal{E}$, representing the transmission links between them, which can be either \gls{isl} or \gls{gsl}, implemented as Radio Frequency (RF) or Free Space Optical (FSO).


\noindent \textbf{Space segment.}
The satellite constellation consists of $N$ satellites evenly distributed across $O$ orbital planes. Each satellite functions as a router and learning agent for \gls{rl}-based solutions. Satellites are positioned at specific \flc{and configurable} altitudes, longitudes, and orbit inclinations, moving according to orbital mechanics and Earth's rotation~\cite{lozanocuadra2024continual}. Satellites move periodically, at the beginning of each time interval, rather than continuously. After a fixed time interval, each satellite is placed in the exact position that it would reach if it had moved continuously during that period. This periodic movement impacts latency calculations by updating transmission and propagation times at each position update. Each satellite has one antenna for \gls{gsl} and four for \gls{isl} (two for inter-plane and two for intra-plane links). Selecting the best \gls{isl} is a dynamic matching problem and consists of establishing the best \glspl{isl} among satellites.
Links are bidirectional, and the network is reconfigured as satellites move, i.e, $\mathcal{G}\flc{_t}$ is built again maintaining previous queue states.


\noindent \textbf{Ground segment.} 
The ground segment consists of a set of configurable ground gateways, which gather the terrestrial traffic from mobile devices. Each gateway aggregates this traffic into large packets for transmission to its nearest satellite, with which it maintains a \gls{gsl}.


\noindent \textbf{Data rate.}
The communication data rate between nodes $i$ and $j$, \flc{$R(i,j)$,} is determined by the highest modulation and coding scheme that ensures reliable communication based on the current \gls{snr}, and zero otherwise, using DVB-S2 technology~\cite{dvb_s2} for realistic data rates assuming free-space loss~\cite{lozanocuadra2024continual}.


\noindent \textbf{Traffic generation.}
We consider a scenario with realistic packet generation, queuing, and transmission, where each gateway transmits data equally split among the other gateways through the \gls{lsatc}, then data is assumed to be distributed to the nearby connected users. The total traffic load $\ell$ in the network is determined by the uplink data generation rate at each gateway and the maximum supported traffic load $\ell$, derived from uplink and downlink rates. The traffic generation follows a Poisson distribution and $\ell$ is configurable by the user. 
\flc{As each gateway sends traffic to each other, the total number of unidirectional flows $U_f$ can be expressed as: $U_f = n_g \cdot (n_g - 1)$, where $n_g$ is the number of active gateways.}


\noindent \textbf{Routing.}
The routing algorithm at each satellite $i$ aims to relay each received packet $p(d)$ towards its destination $d$. Each satellite has a transmission buffer with a maximum capacity of $Q^{\max}$, operating under a \gls{fifo} strategy. If the buffer is not empty, the satellite takes the Head of Line packet and delivers it to one of its linked nodes following the chosen routing policy. Any packet arriving at a full buffer is dropped.


\noindent \textbf{Latency.}
The one-hop latency to transmit a packet from $i$ to $j$ depends on three factors: queue time, transmission time, and propagation time~\cite{lozanocuadra2024continual}. The queue time at the transmission queue is the elapsed time since the packet is ready to be transmitted until the beginning of its transmission. The transmission time is the time taken to transmit the packet based on the transmission rate. The propagation time is the time it takes for the signal to travel the distance between $i$ and $j$, $||ij||$. This latency model considers varying traffic loads, where propagation time is significant in low traffic but queue time increases under high traffic conditions~\cite{Rabjerg2021}.




\begin{figure}[t]
   \centering
   \includegraphics[width=.95\columnwidth]{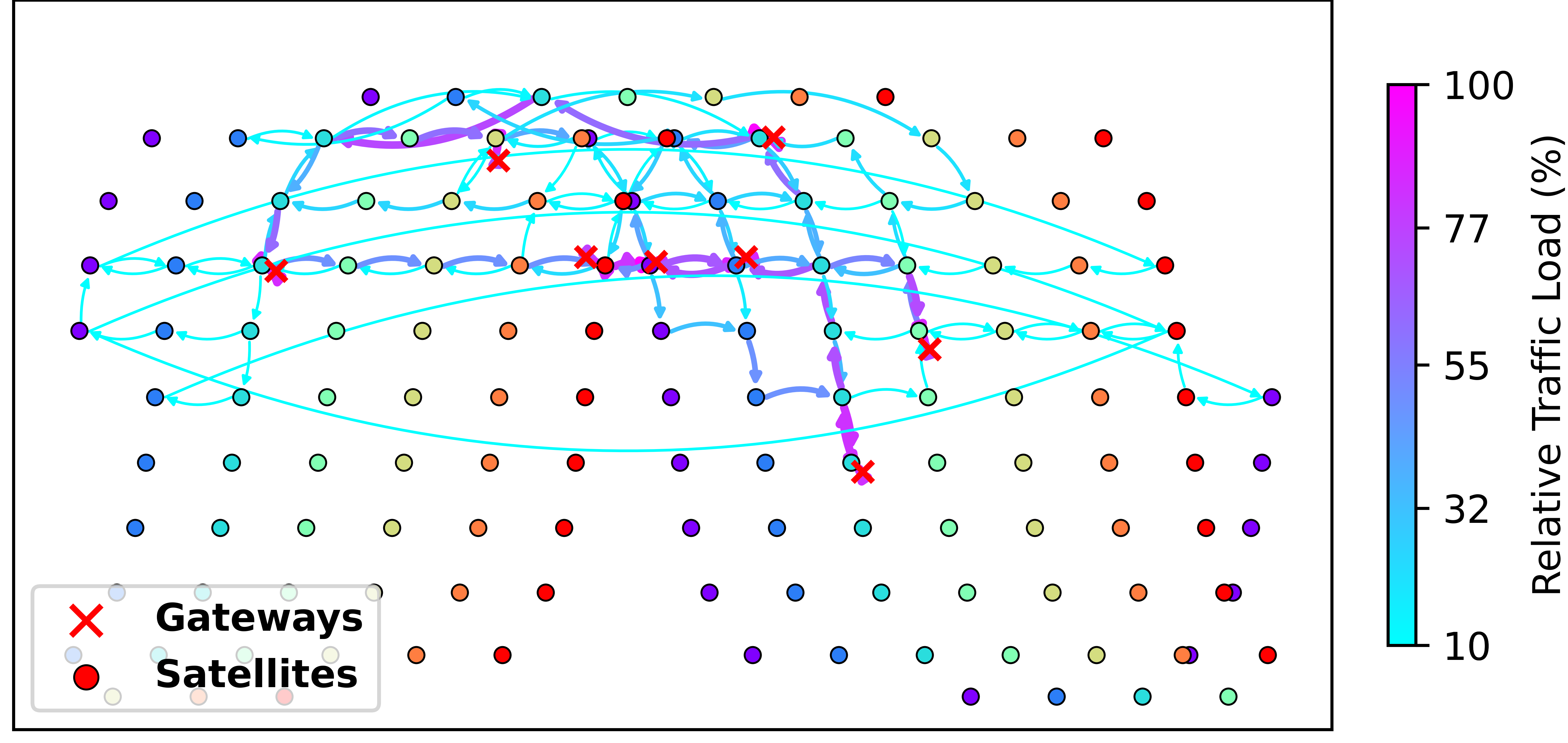}
   \caption{\gls{ma-drl}'s exploitation phase congestion test for all routes output with 8 active gateways.}
   \label{fig:Paths}\vspace{-0.5cm}
\end{figure}


\section{Routing algorithms}

Different routing policies are implemented in the simulator. \flc{On one hand}, we have the deterministic ones, all of them based on shortest path Dijkstra's algorithm~\cite{dijkstra1959}, where the edge weights are minimized in centrally with full knowledge of the constellation. Each method minimizes a different weight: (1) \textbf{Data Rate}, where the edge weights between two nodes $i$ and $j$ are determined by the inverse of the data rate between nodes, namely $w_{i,j}=1/R(i,j)$. This is a traditional routing approach that leads to choosing routes with high data rate links; (2) \textbf{Slant Range}, where the edge weights between $i$ and $j$ nodes are defined by the distance between them, $||ij||$, in order to minimize propagation times, and the (3) \textbf{Hop}, where all edges have the same weight, 1, where the total number of jumps is minimized.

On the other hand, other two \gls{rl}-based routing policies are implemented, specifically the ones developed in our previous work. \flc{Firstly}, the \textbf{Q-Routing} policy, developed in~\cite{soret2023qlearning}. Q-Tables are created automatically with NumPy~\cite{van2011numpy} to increase efficiency. They will store the learnt knowledge during the training process. The user can choose if it wants the algorithm to explore and make random routing actions or import pre-trained Q-Tables and exploit its knowledge to use them \flc{as} routing policy.

\noindent Secondly, \textbf{\gls{ma-drl}} from~\cite{lozanocuadra2024multiagent,lozanocuadra2024continual} is implemented. The \glspl{dnn} are initialized and trained with \emph{Keras}~\cite{ketkar2017introduction}. \gls{ddqn} ~\cite{van2016deep} is implemented and its usage is configurable. It is also possible either to import pre-trained \glspl{dnn} or not and choose between the offline and the online phase of the algorithm.



\begin{figure}[t]
   \centering
   \includegraphics[width=.95\columnwidth]{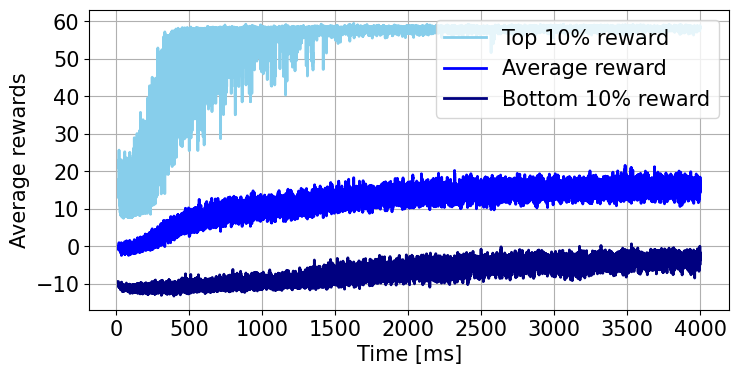}
   \caption{Rewards over time of the offline phase of \gls{ma-drl} with 8 active gateways. The highest rewards are given after a packet has been delivered to the receiving gateway.}
   \label{fig:8GTsRewards}\vspace{-0.5cm}
\end{figure}


\section{Setup and general settings}

The simulator, running in \textbf{Python 3.9}, is multi-platform and has been tested on Windows, Linux, and Mac systems. The user can install the required packages listed in the \emph{requirements.txt} file using pip. It is advisable to create a virtual environment or an Anaconda environment for better management.

The simulator is highly configurable, allowing users to adjust various parameters to suit their specific needs, as illustrated in Fig.~\ref{fig:system}. Key configurable parameters include: (a) \textbf{Routing Policy}, including the shortest path-based, where the Data Rate, Slant Range or Hop can be set as weights, and the \gls{rl}-based options; (b) \textbf{Ground Segment} settings, such as the number and locations of active gateways, as well as traffic generation $\ell$; (c) \textbf{Space Segment} parameters, which cover constellation design (configurable elements include the number of orbital planes and its inclination angle, satellites per plane, and the choice between Walker delta and Walker star designs), \gls{isl} matching (Greedy or Markovian\flc{~\cite{Leyva-Mayorga2021}}), and orbital motion (which can be sped up or slowed down); (d) \textbf{Learning Hyperparameters}, including rewards and penalties, exploration $\epsilon$, learning $\alpha$ and gamma $\gamma$ rates, state preprocessing, and training modes (Import pre-trained models for \gls{rl}-based policies and choosing between online and offline phases for \gls{ma-drl}); and (e) \textbf{Communication Setup}, such as physical constants, uplink and downlink parameters, and packet size. Additionally, the user can configure the simulator to plot the environment every time the constellation moves to visualize system motion, as in Fig.~\ref{fig:ISLs_Kepler}, and to plot the path of each delivered packet over this to track its journey through the network.

With all these settings configured, the simulator is now ready to run simulations and generate results.


\begin{figure}[t]
   \centering
   \includegraphics[width=.95\columnwidth]{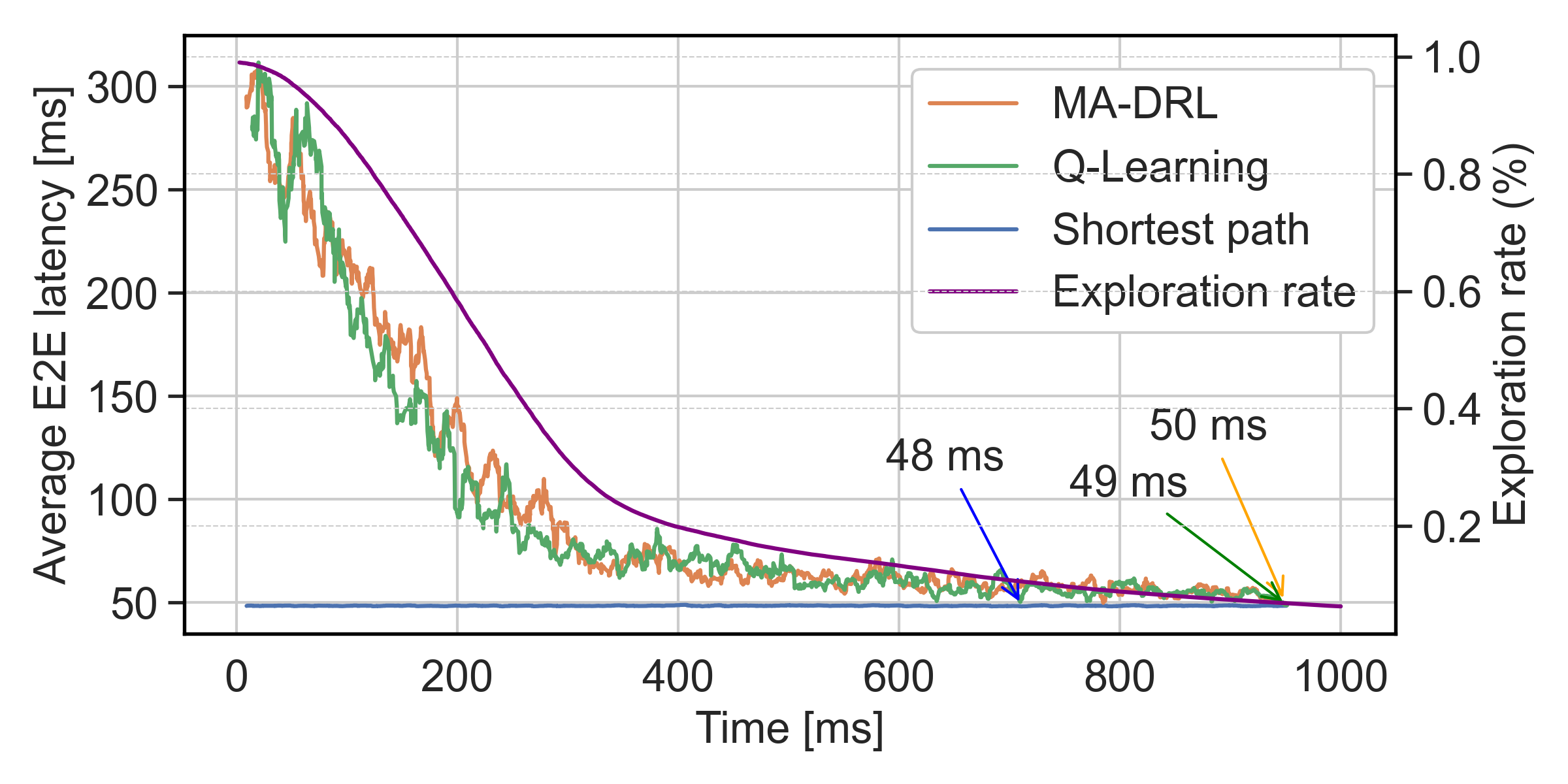}
   \caption{\flc{\gls{e2e} latency vs time vs $\epsilon$ connecting one gateway in Malaga, Spain and another one in Los Angeles, USA, through the Starlink constellation during the offline phase of the \gls{rl}-based methods. It can be appreciated how both methods learn to find the optimal path in less than 1 second.}}
   \label{fig:2GTsOffline}\vspace{-0.5cm}
\end{figure}


\section{Results and analysis}

The default ground segment has up to $18$ active (transmitting and receiving) gateways distributed across the Earth, mainly following KSAT's deployment\footnote{https://www.ksat.no/services/ground-station-services/}, but more gateways can be added easily.

Moreover, four real constellations are implemented: (1) \emph{Kepler} constellation design, with $O=7$ orbital planes at heights $h = 600$ km and $N = 20$ satellites per orbital plane, as illustrated in Fig.~\ref{fig:ISLs_Kepler}; (2) \emph{Iridium Next} constellation, with $O=6$, $h=780$ km and $N=11$; (3) the \emph{OneWeb} constellation, with $O=36$, $h=1200$ km and $N=18$; and (4) \emph{Starlink} orbital shell at $h=550$ km, with $O=72$ and $N=22$. The three first constellations follow a \emph{Walker star} architecture, while the Starlink shell follows a \emph{Walker delta} architecture~\cite{leyva2022ngso}. Moreover, two additional artificial constellations are implemented for testing. Additionally, two \gls{isl} matching algorithms are implemented: (1) The Markovian solution proposed in~\cite{Leyva-Mayorga2021} and (2) a \emph{Greedy} approach, where each satellite connects with immediate neighbors within its plane and closest counterparts in adjacent planes in both East and West directions, optimizing for latency and data rates, as shown in Fig.~\ref{fig:ISLs_Kepler}.

When a simulation ends, it automatically outputs  a set of results in the form of figures and text files (Fig.~\ref{fig:system}). Within the \textbf{figures}, a map like Fig.~\ref{fig:ISLs_Kepler} with the system model information is saved. An update of this figure is also saved as the constellation moves if desired. Then, a congestion test per route and for all routes between gateways is done, in order to see what nodes and edges did the packets went through, as shown in Fig.~\ref{fig:Paths}. If one of the \gls{rl}-based routing policies is chosen, one figure with the exploration rate $\epsilon$ and training stamps and another one with the received rewards (Fig.~\ref{fig:8GTsRewards}) are also saved. Other figures saved are related to the average \gls{e2e} latency vs time vs $\epsilon$, \flc{similar to Fig.~\ref{fig:2GTsOffline}, but with just one routing policy, and to the} queue lengths. On the other hand, the output \textbf{files} include several \emph{.csv} with extensive information about each packet's \flc{path} and its latency, rewards, exploration rates, training stamps, hyper-parameters and a \emph{.txt} log-file, that saves everything that happened during the simulation and gives some statistics like like the latency broken down by average queue, transmission and propagation times, packets delivered vs stuck and/or lost, most used links, etc. Lastly, if either the \gls{ma-drl} or the Q-Routing algorithm was chosen for routing, the trained \glspl{dnn} (57Kb for the Q-Network and 27 Kb for the Q-Target) or Q-Tables (21Kb for 8 active gateways) are saved, respectively.


\begin{figure}[t]
   \centering
   \includegraphics[width=.95\columnwidth]{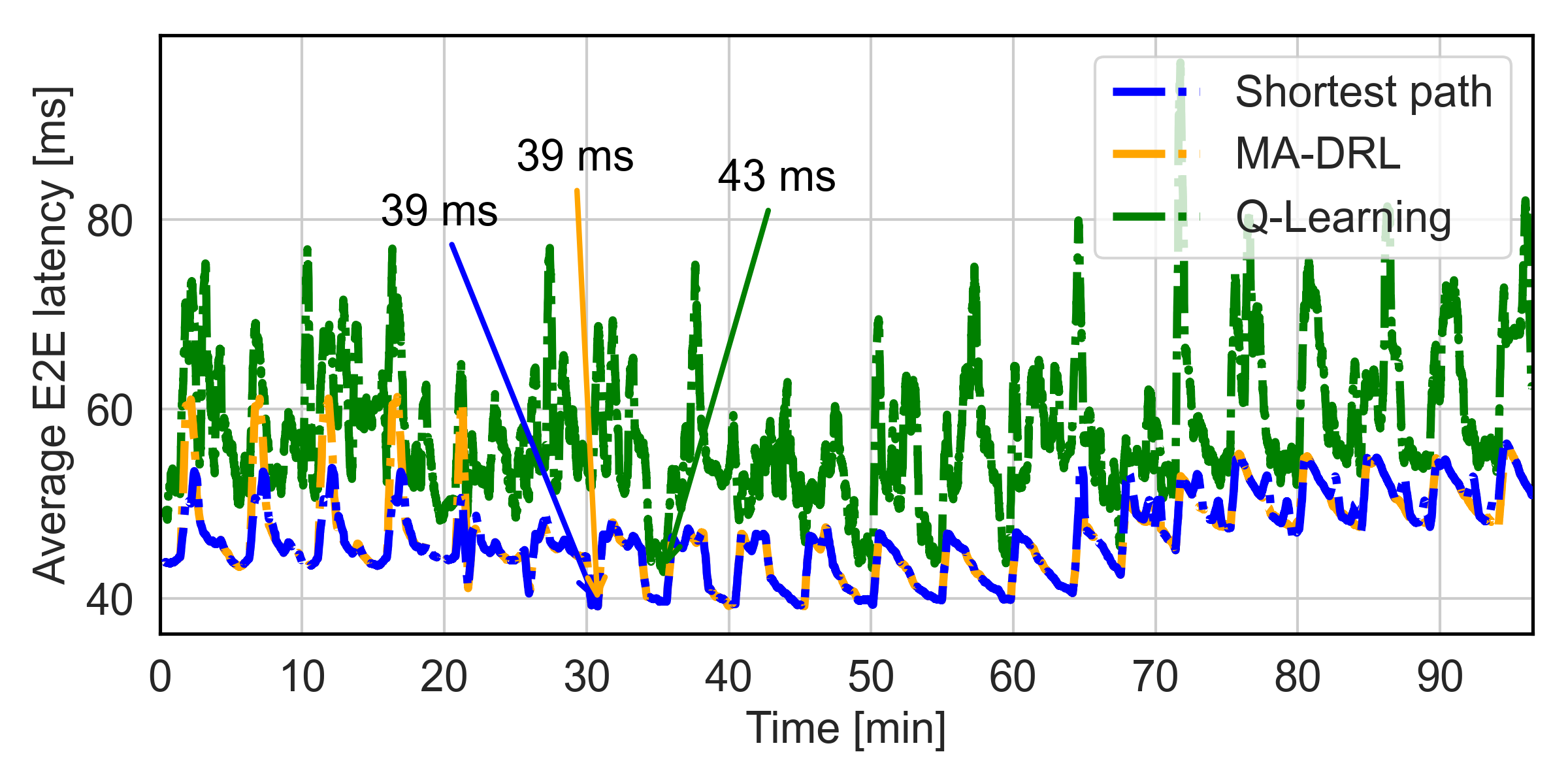}
   \caption{Average \gls{e2e} latency over an orbital period. \flc{The fluctuations are given by the movement of the satellites and the resulting changes in the routed followed by the packets.}
   }
   \label{fig:movement}\vspace{-0.5cm}
\end{figure}

\begin{figure}[t]
   \centering
   \includegraphics[width=.95\columnwidth]{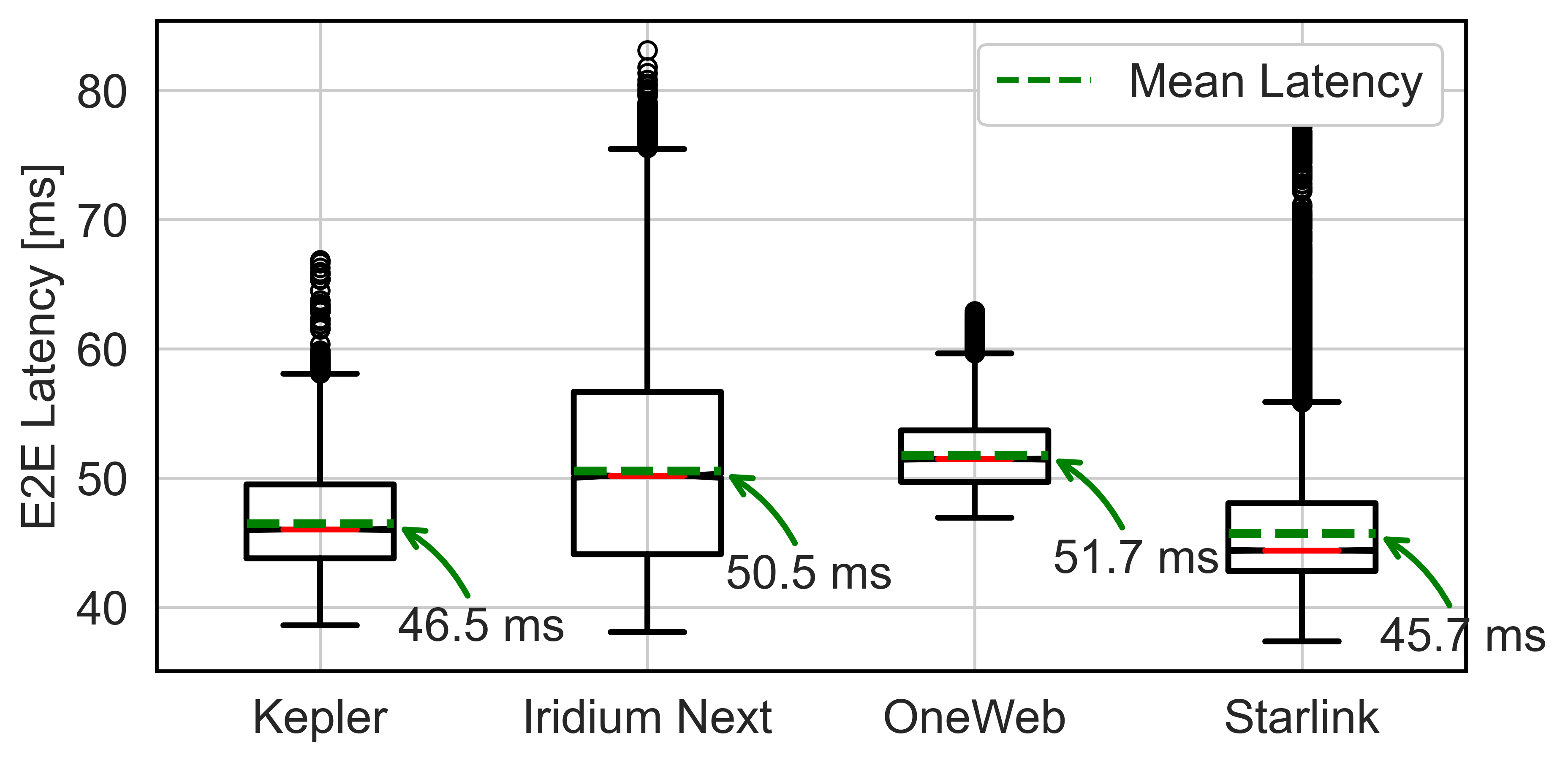}
   \caption{Box plot of the \gls{e2e} latency of the four constellation topologies with the \emph{Shortest Path} policy {after one orbital period is completed}.
   }
   \label{fig:boxPlot}\vspace{-0.5cm}
\end{figure}

\flc{In the \emph{Jupyter notebook}, we conduct further \textbf{post-processing} analysis and explore more complex results.}
\flc{A comparison between the \flc{\emph{Shortest path}} routing policy, Q-Routing~\cite{soret2023qlearning} and \gls{ma-drl}~\cite{lozanocuadra2024multiagent,lozanocuadra2024continual} at their offline phase is shown in Fig.~\ref{fig:2GTsOffline}.}
Additionally, a \flc{dynamic} comparison \flc{of these policies} at their online phase is shown in Fig.~\ref{fig:movement}, where the constellation has moved to complete one orbital period in $96$ minutes, with the satellite positions being updated at intervals of $15$ seconds. 
\flc{Notably, even with only partial knowledge of the constellation, \gls{ma-drl} consistently maintains the baseline latency obtained with the \flc{\emph{Shortest path}} policy, which has full knowledge of the constellation.}
Moreover, we elaborate on the comparison of the four architectures in Fig.~\ref{fig:boxPlot}, where the distribution of the \gls{e2e} latency is depicted in a box plot {when the \emph{Shortest path} is applied} \flc{among one orbital period too}. 
We observe that \emph{Kepler} and \emph{Starlink} obtain the smallest average latency, although the latter presents {more outliers}. 
This figure helps to illustrate the behavior of the constellations and highlights the usage of the simulator to test different constellation architectures.


\flc{Additionally, as in \gls{ma-drl}, each satellite is an independent agent during the online phase, we conducted a \gls{cka}~\cite{kornblith2019similarity} analysis to compare the differences between each agent's local model after 1 second with varying traffic patterns around the globe. Each satellite learns and adapts its routing decisions based on these traffic patterns, resulting in distinct updates to their local models. Consequently, these models exhibit differences. To homogenize the models, we applied post-processing \gls{sfl} techniques: Initially among neighboring satellites, Model Anticipation; then, among orbital planes, Orbital Plane Aggregation (\gls{sfl}); and finally, across the entire constellation, Full Aggregation (\gls{sfl})~\cite{lozanocuadra2024continual}. 
}


\section{Conclusion and Future work}

The development of an open source \gls{ma-drl} simulator for satellite network routing provides a robust platform for testing and implementing various routing algorithms \flc{in \emph{Python}, where different machine learning libraries can be leveraged}. The simulator’s high configurability and realism allows for comprehensive evaluation of different constellation designs and communication setups. The results highlight the effectiveness of \gls{rl}-based routing policies compared to traditional methods, demonstrating significant improvements in \flc{\gls{e2e}} latency and overall network performance.

\flc{\textbf{Future directions} include: (1) Developing an \gls{sfl} framework to enable aggregation during the online phase of \gls{ma-drl} rather than implementing it as a post-processing analysis; (2) implementing a two-tier mesh network for UE-satellite-UE communications, enabling ground moving users to connect directly to satellites without the need for gateways; (3) increase the complexity of satellites with regenerative capabilities; and (4) implementing different types of traffic with splittable flows.}






\printbibliography
\addcontentsline{toc}{section}{References}

\end{document}